\pdfoutput=1

\documentclass[11pt]{article}

\usepackage{EMNLP2023}

\usepackage{times}
\usepackage{latexsym}

\usepackage{tipa}
\usepackage{amsmath}
\usepackage{amssymb}

\providecommand{\tightlist}{%
  \setlength{\itemsep}{0pt}\setlength{\parskip}{0pt}}

\usepackage{cleveref}
\creflabelformat{equation}{#2\textup{#1}#3} 
\usepackage{graphicx}
\usepackage{subcaption}
\usepackage{placeins}
\usepackage{booktabs}

\usepackage{comment}

\newcommand\blfootnote[1]{%
  \begingroup
  \renewcommand\thefootnote{}\footnote{#1}%
  \addtocounter{footnote}{-1}%
  \endgroup
}

\title{The neural dynamics of auditory word recognition and integration}

\author{Jon Gauthier$^{1,2}$ \and Roger Levy$^1$\\
$^1$ Department of Brain and Cognitive Sciences,
Massachusetts Institute of Technology \\
$^2$ Department of Neurological Surgery, University of California San Francisco \\
\href{mailto:jon@gauthiers.net}{\tt jon@gauthiers.net}, \href{mailto:rplevy@mit.edu}{\tt rplevy@mit.edu}}

\begin{document}

\maketitle

\begin{abstract}
Listeners recognize and integrate words in rapid and noisy
everyday speech by combining expectations about upcoming content with 
incremental sensory evidence. We present a computational model of word recognition which
formalizes this perceptual process in Bayesian decision theory. We fit
this model to explain scalp EEG signals recorded as subjects passively
listened to a fictional story, revealing both the dynamics of the online
auditory word recognition process and the neural correlates of the recognition
and integration of words.

The model reveals distinct neural processing
of words depending on whether or not they can be quickly recognized. While all words trigger a neural response characteristic of probabilistic integration --- voltage modulations predicted by a word's surprisal in context --- these modulations are amplified for words which require more than roughly 150 ms of input to be recognized. We observe no difference in
the latency of these neural responses according to words' recognition
times.
Our results are consistent with a two-part model of speech comprehension, combining an eager and rapid process of word recognition with a temporally independent process of word integration.
However, we also developed alternative models of the scalp EEG signal not incorporating word recognition dynamics which showed similar performance improvements.
We discuss potential future modeling steps which may help to separate these hypotheses.
\end{abstract}

\blfootnote{Code to reproduce our analyses is available at \href{https://github.com/hans/word-recognition-and-integration}{\texttt{\scriptsize{}github.com/hans/word-recognition-and-integration}}.}
Psycholinguistic studies at the neural
and behavioral levels have detailed how listeners actively predict
upcoming content at many levels of linguistic representation \citep{kuperberg2016what},
and use these predictions to drive their behavior far before the relevant
linguistic input is complete \citep{allopenna1998tracking}.
One well-studied neural correlate of this prediction-driven comprehension
process is the N400 ERP, a centro-parietally distributed negative voltage
modulation measured at the scalp by electroencephalogram (EEG) which peaks
around 400 ms after the onset of a word. This negative component is amplified for words which
are semantically incompatible with their sentence or discourse context
\citep{kutas1984brain,brown1993processing,kutas2011thirty,heilbron2022hierarchy}.
This effect has been taken as evidence that comprehenders actively predict
features of upcoming words \citep{delong2005probabilistic,kuperberg2016what,kuperberg2020tale}. On one popular
account, predictions about upcoming content are used to pre-activate
linguistic representations likely to be used when that content arrives. The
N400 reflects the integration of a recognized word with its context, and
this integration is facilitated just when the computational paths taken by the
integration process align with those already pre-activated by the listener \citep{kutas2011thirty,federmeier2007thinking}.

Despite the extensive research on the N400 and its computational interpretation, its relationship with the upstream process of word recognition is still not well understood.
Some authors have argued that
integration processes should be temporally yoked to word recognition: that is,
comprehenders should continue gathering acoustic evidence as to the identity
of a word until they are sufficiently confident to proceed with subsequent integration
processes \citep{marslen1987functional}.
It is also possible, however, that integration processes are insensitive to
the progress of word recognition: that integration is a temporally regular
semantic operation which begins regardless of the listener's confidence
about the word being spoken \citep{hagoort2008fractionation,federmeier2009time}.

Experimental studies have attempted to assess the link between these
two processes, modeling the timing of word recognition
through an offline behavioral paradigm known as \emph{gating} 
\citep{grosjean1980spoken}: by presenting incrementally longer clips
of speech to subjects and asking them to predict what word is being
spoken, authors estimate the time point at which there is sufficient
information to identify a word from its acoustic form. Several EEG studies have asked whether the N400 response varies with respect to
this estimate of word recognition time, but have arrived at contradictory answers to this question \citep{vandenbrink2006cascaded,o2002electrophysiological}.

In this paper, we introduce a computational model which targets these
dynamics of word recognition, and their manifestation in neural EEG
signals recorded during naturalistic listening. The model allows us
to connect trial-level variation in word recognition times to aspects of the
neural response to words. We use the model to address two cross-cutting 
questions:
\begin{itemize}
    \item\textbf{Onset:} Are words integrated only after they are successfully recognized,
    or is the timing of integration insensitive to the state of word recognition?
    \item\textbf{Response properties:} Does the shape of the neural response to words
    differ based on their recognition times? If so, this could indicate
    distinct inferential mechanisms deployed for words depending on their
    ease of recognition.
\end{itemize}
We jointly optimize the cognitive and neural parameters of this model to explain
EEG data recorded as subjects listened to naturalistic English speech.
Model comparison results suggest that semantic integration processes are \underline{not} temporally yoked to the
status of word recognition: the neural traces of word integration have
just the same temporal structure, regardless of when words are successfully recognized. 
However, the neural correlates of word integration qualitatively 
differ based on the status of word recognition: words not yet recognized by the 
onset of word integration exhibit significantly different
neural responses.

These results suggest a two-part model of word recognition and integration. First, the success of our word recognition model in predicting the neural response to words suggests that there exists a rapid lexical interpretation process which integrates prior expectations and acoustic evidence in order to pre-activate specific lexical items in memory. Second, an independent integration process composes these memory contents with a model of the context, following a clock which is \emph{insensitive} to the specific state of word recognition.

It is necessary to moderate these conclusions, however: we also develop alternative models of the neural correlates of word integration which improve beyond the performance of our baselines, without incorporating facts about the dynamics of word recognition. We discuss in \Cref{sec:discussion} how more elaborate neural linking theories will be necessary to better separate these very different cognitive pictures of the process of word recognition and its neural correlates.

\section{Model}
\label{sec:model}

\newcommand\recpoint{\ensuremath{k_i^\ast}}
\newcommand\rectime{\ensuremath{\tau_i}}
\newcommand\scatterpoint{\ensuremath{\alpha}}
\newcommand\priorscatterpoint{\ensuremath{\alpha_p}}
\newcommand\threshold{\ensuremath{\gamma}}
\newcommand\temperature{\ensuremath{\lambda}}

Our model consists of two interdependent parts: a cognitive model of the
dynamics of word recognition, and a neural model that estimates how these
dynamics drive the EEG response to words.

\begin{table}[t]
    \centering
    \resizebox{\linewidth}{!}{
    \begin{tabular}{r|p{5.75cm}l}
        \toprule
         & Meaning & Bounds \\
        \midrule
        
        \threshold & Recognition threshold
        (\cref{eqn:recognition-point}) & $(0, 1)$ \\
        
        \temperature & Evidence temperature
        (\cref{eqn:likelihood}) & $(0, \infty)$ \\
        
        \scatterpoint & Scatter point
        (\cref{eqn:recognition-time}) & $(0, 1)$ \\
        
        \priorscatterpoint & Prior scatter point
        (\cref{eqn:recognition-time}) & $(0, 1)$ \\
        
        \midrule
        \recpoint & Word $w_i$'s recognition point
        (\cref{eqn:recognition-point}) & $\{0, 1, \dots, |w_i|\}$ \\
        \rectime & Word $w_i$'s recognition time (\cref{eqn:recognition-time}) & $[0, \infty)$ \\
        \bottomrule
    \end{tabular}
    }
    \caption{Cognitive model parameters and outputs.}
    \label{tbl:cognitive-parameters}
\end{table}

\subsection{Cognitive model}

We first design a cognitive model of the dynamics of word recognition in context, capturing how a listener forms incremental beliefs about the word they are hearing $w_i$ as a function of the linguistic context $C$ and some partial acoustic evidence $I_{\le k}$. We formalize this as a
Bayesian posterior \citep{norris2008shortlist}:
\begin{equation}
    P(w_{i} \mid C,I_{\leq k}) \propto P(w_{i} \mid C)\; P(I_{\leq k} \mid w_{i}) \label{eqn:posterior}
\end{equation}
which factorizes into a prior expectation of the word {\(w_{i}\)} in context
(first term) and a likelihood of the partial evidence of
{\(k\)} phonemes {\(I_{\leq k}\)} (second term).
This model thus asserts that the context $C$ and the acoustic
input $I_{\leq k}$ are conditionally independent given $w_i$.
We parameterize the
prior {\(P(w_{i} \mid C) = P(w_{i} \mid w_{< i})\)} using a left-to-right neural network
language model. The likelihood is a noisy-channel phoneme recognition model:
\begin{equation}
    P(I_{\leq k} \mid w_{i}) \propto \prod\limits_{1 \leq j \leq k}P(I_{j} \mid w_{ij})^{\frac{1}{\temperature}} \label{eqn:likelihood}
\end{equation}
where
per-phoneme confusion probabilities are drawn from prior phoneme
recognition studies \citep{weber2003consonant} and reweighted by a
temperature parameter \temperature.

We evaluate this posterior for every word with each incremental phoneme, from $k=0$ (no input) to $k=|w_i|$ (conditioning on all of the word's phonemes). We define a hypothetical
cognitive event of \emph{word recognition} which is time-locked to the phoneme
\recpoint{} where this posterior first exceeds a confidence threshold
\threshold:
\begin{equation}
    \recpoint = \min_{0 \leq k \leq |w_i|} \left\{k \mid P(w_{i} \mid C,I_{\leq k}) > \threshold \right\} \label{eqn:recognition-point}
\end{equation}
We define a word's \emph{recognition time} \rectime{} to be a fraction \scatterpoint{}
of the span of the \recpoint-ith phoneme. In the special
case where $\recpoint=0$ and the word is confidently identified prior to acoustic
input, we take \rectime{} to be a fraction \priorscatterpoint{} of its first phoneme's
duration (visualized in \Cref{fig:scatter-logic}):
\begin{equation}
    \rectime = \left\{ \begin{matrix}
{\text{ons}_{i}(\recpoint) + \scatterpoint\, \text{dur}_{i}(\recpoint)} & {\text{if~}{\recpoint > 0}} \\
{\priorscatterpoint\, \text{dur}_{i}(1)} & {\text{if~}{\recpoint = 0}}
\end{matrix} \right.
    \label{eqn:recognition-time}
\end{equation}
\noindent where $\text{ons}_i(k)$ and $\text{dur}_i(k)$ are the onset time (relative to word onset) and duration of the $k$-th phoneme of word $i$, and $\scatterpoint,\priorscatterpoint$ are free parameters fitted jointly with
the rest of the model.

\subsection{Neural model}
\label{sec:model-neural}

We next define a set of candidate linking models
which describe how the dynamics of the cognitive model (specifically,
word recognition times $\tau_i$) affect observed neural responses. These
models are all variants of a temporal receptive field model \citep[TRF;][]{lalor2009resolving,crosse2016multivariate},
which predicts scalp EEG data over $S$ sensors and $T$ samples, $Y \in \mathbb R^{S \times T}$, as a convolved set of linear responses to lagged features of the stimulus:
\begin{equation}
    Y_{st} = \sum_f \sum_{\Delta=0}^{\tau_f} \Theta_{f,s,\Delta} \times \mathbf X_{f, t-\Delta} + \epsilon_{st}
    \label{eqn:trf}
\end{equation}
\noindent where $\tau_f$ is the maximum expected lag (in seconds) between the onset of a feature $f$ and its correlates in the neural signal; and the inner sum is accumulated in steps of the relevant neural sampling rate. This deconvolutional model estimates a characteristic linear response linking each feature of the stimulus to the neural data over time.
The model allows us to effectively uncover the neural response to individual stimulus features in naturalistic data, where stimuli (words) arrive at a fast rate, and their neural responses are likely highly convolved as a consequence \citep{crosse2016multivariate}.

We define a feature time series $X_t \in \mathbb R^{d_t \times T}$ containing $d_t$ features of the objective auditory stimulus, such as
acoustic and spectral features, resampled to match the $T$ samples of the neural time series.
We also define a word-level feature matrix $X_v \in \mathbb R^{d_w \times n_w}$ for the $n_w$ words in the stimulus. Crucially, $X_v$ contains estimates of each word's surprisal (negative log-probability) in context. Prior studies suggest that surprisal indexes the peak
amplitude of the naturalistic N400 
\citep{frank2015erp,gillis2021neural,heilbron2022hierarchy}.

We assume that
{\(X_{t}\)} causes a neural response independent of word recognition
dynamics, while the neural response to features {\(X_{v}\)} may vary
as a function of recognition dynamics.
These two feature matrices will be merged together
to yield the design matrix $\mathbf X$ in \Cref{eqn:trf}.

\begin{table}[t]
    \centering
    \resizebox{\linewidth}{!}{
    \begin{tabular}{l|lp{4.5cm}}
        \toprule
        Model name & Onset & Response properties \\
        \midrule
        Baseline & 0 & unitary linear response \\
        Shift & $\rectime$ (\cref{eqn:recognition-time}) & unitary linear response \\
        Variable & 0 & independent linear responses for early\nobreakdash-, mid\nobreakdash-, and late\nobreakdash-recognized words \\
        Prior-variable & 0 & independent linear responses for low\nobreakdash-, mid\nobreakdash-, and high-surprisal words \\ 
        \bottomrule
    \end{tabular}
    }
    \caption{Neural linking models with different commitments about the temporal onset of word features (relative to word onset) and the flexibility of the parameters linking word features to neural response.}
    \label{tbl:models}
\end{table}

\begin{figure}
    \begin{subfigure}{\linewidth}
        \centering
        \includegraphics[width=\linewidth]{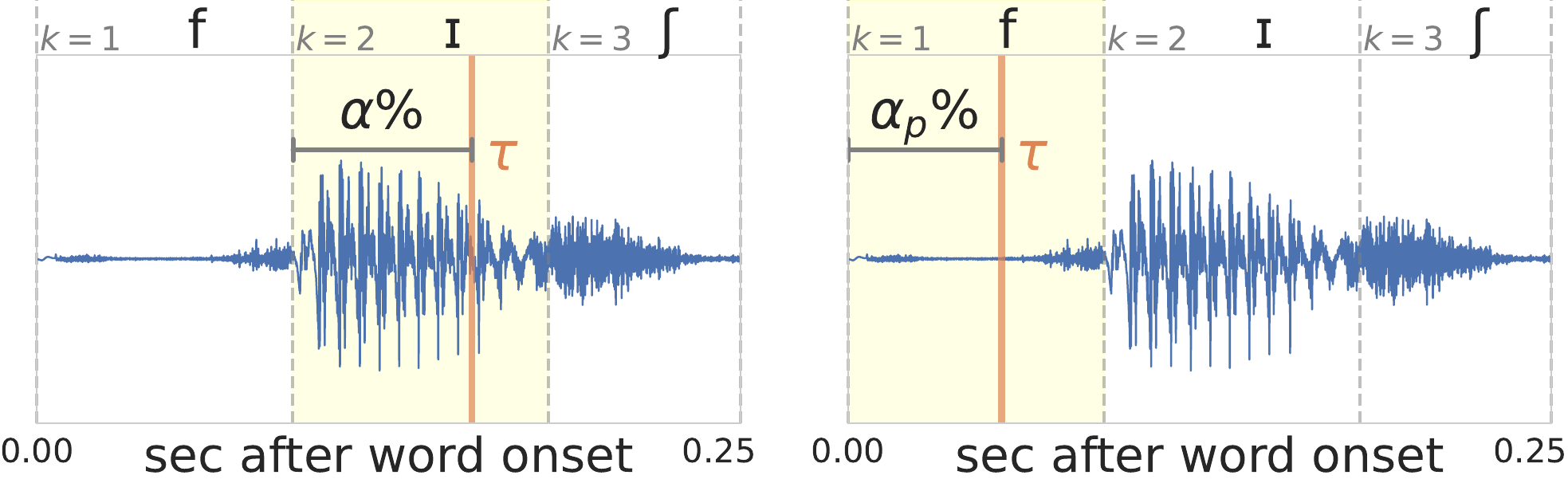}
        \caption{Computation of recognition time \rectime{} for a recognition point after phoneme $\recpoint = 2$ (left) or recognition prior to input, $\recpoint = 0$ (right) for a spoken word \emph{fish} \textipa{/fIS/}. See \cref{eqn:recognition-time}.}
        \label{fig:scatter-logic}
    \end{subfigure}
    \begin{subfigure}{\linewidth}
        \centering
        \includegraphics[width=0.85\linewidth]{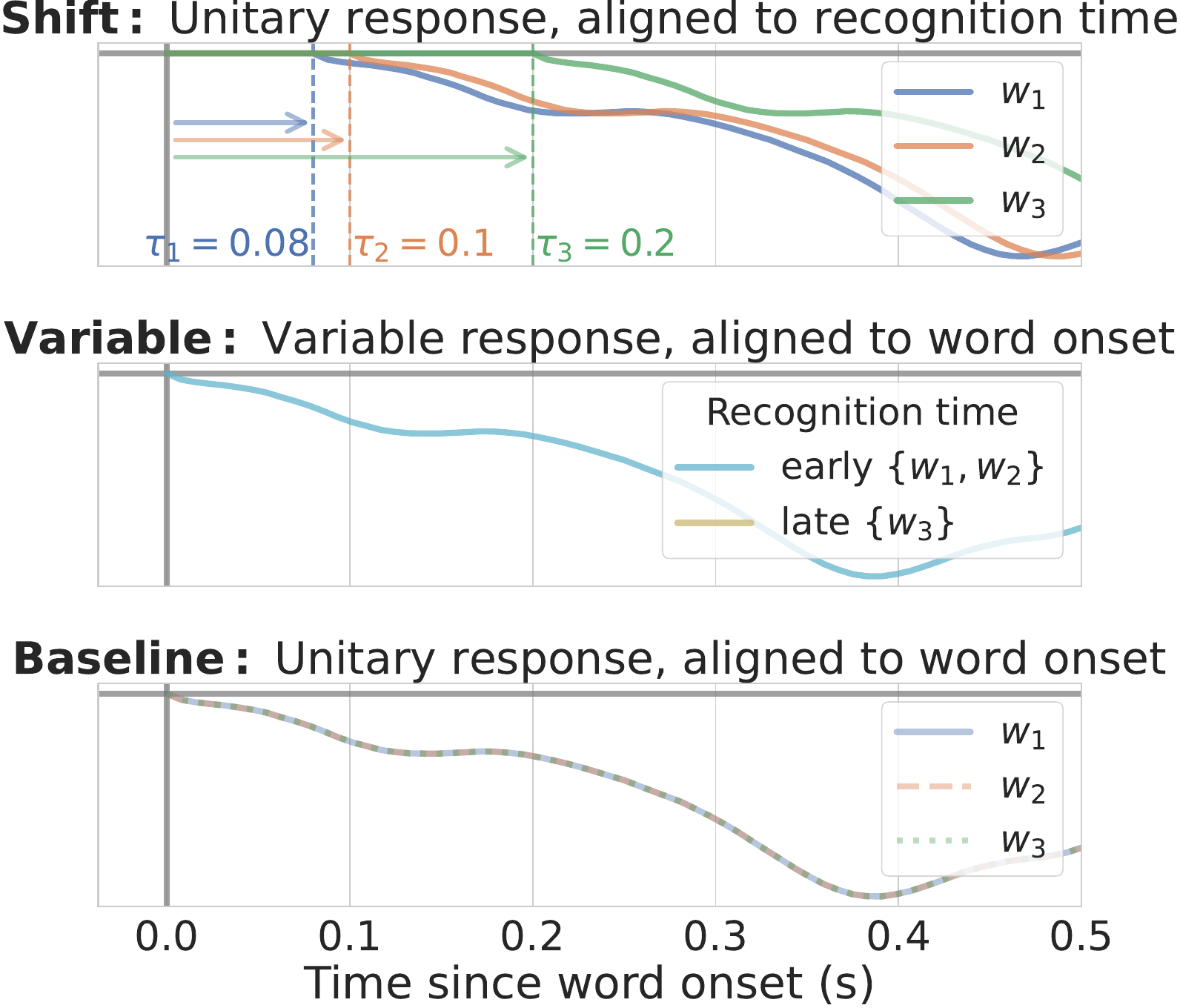}
        \caption{Candidate neural model logic linking three words' recognition times $\tau_i$ to neural modulations by surprisal.}
        \label{fig:neural-linking-models}
    \end{subfigure}
    \caption{Sketches of model logic.}
\end{figure}


We enumerate several possible classes of neural models which describe different
ways that a word's recognition time \rectime{} may affect the neural response. Each
model class constitutes a different answer to our framing questions of \textbf{onset}
and \textbf{response properties} (\Cref{tbl:models} and \Cref{fig:neural-linking-models}), by specifying different featurizations of word-level properties $X_v$ in the TRF design matrix $\mathbf X$:
\begin{enumerate}
    \item
        Unitary response aligned to word onset (\emph{baseline model}): All words exhibit a unitary linear neural response to recognition and integration, time-locked to the word's onset in the stimulus. This baseline model, which does not incorporate the cognitive dynamics of recognition in any way, is what has been assumed by prior naturalistic modeling work.

        This model asserts that each word's features $X_{vi}$ trigger a neural response beginning at the onset of word $i$, and that this neural response can be captured by a single characteristic response to all words.

    \item
        Unitary response aligned to recognition time (\emph{shift model}): All words exhibit a unitary linear neural response to recognition and integration, time-locked to the word's recognition time \rectime{}.

        This model asserts that each word's features $X_{vi}$ trigger a neural response beginning at $\tau_i$ seconds after word onset, and that this neural response can be captured by a single characteristic response to all words.
        
    \item
        Variable response by recognition time, aligned to word onset (\emph{variable model}): Words exhibit a differential neural response to recognition and integration based on their recognition time. The temporal onset of these integration processes is insensitive to the progress of word recognition.

        We account for variable responses by defining a quantile split $Q : \tau \to \mathbb N$ on the inferred
        recognition times \rectime. We then estimate distinct TRF parameters for the features of words in each quantile.
        
        This model thus asserts that it is possible to group words by their
        recognition dynamics such that they have a characteristic neural response
        within-group,
        but differ freely between groups.

    \item 
        Variable response by word surprisal, aligned to word onset (\emph{prior-variable model}): This model is identical to the above \emph{variable model}, except that words are divided into quantiles based on their surprisal in context rather than their recognition time.

        This model instantiates the hypothesis that the 
        shape of the neural response to words varies based on listeners'
        expectations, but only those driven by the
        preceding linguistic context. On this reading, words are
        pre-activated according to their \emph{prior} probability, rather than their rapidly changing \emph{posterior}
        probability under some acoustic input.\footnote{This reading
        is compatible with pre-activation theories \citep[e.g.][]{brothers2021word}. At their present level of specificity, it is unclear whether this focus on prior probability is a substantive
        commitment, or simply a choice of modeling expediency.}

\end{enumerate}

For a set of recognition time predictions {\(\tau_{i}\)}, we estimate
within-subject TRFs under each of these linking models, yielding
per-subject parameters $\Theta_j$,
describing the combined neural response to objective stimulus features
and word-level features. This estimation procedure allows for
within-subject variation in the shape of the neural response.

\section{Methods and dataset}
\label{sec:dataset}

We jointly infer\footnote{We conduct tree-structured Parzen estimator random search \citep{bergstra2011algorithms} with Optuna \citep{optuna_2019}.} across-subject parameters of the cognitive model (\Cref{tbl:cognitive-parameters}) and within-subject
parameters of the neural model in order to minimize regularized L2 loss on EEG
data, estimated by 4-fold cross-validation. We then compare the fit
models on held-out test data, containing 25\% of the neural time series data for each
subject. For each comparison of models $m_1, m_2$, we compute the Pearson correlation 
coefficient $r$ between the predicted and observed neural response for each subject 
at each EEG sensor $s$. We then use paired $t$-tests to ask whether the within-subject difference in $r$ pooled
across sensors significantly differs between $m_1$ and $m_2$:
\begin{equation}
    \frac 1 S \sum_{s=1}^S r\left( Y_s, \hat Y_{m_1,s} \right) \stackrel{?}{>} \frac 1 S \sum_{s=1}^S r\left( Y_s, \hat Y_{m_2,s} \right)
\end{equation}

\paragraph{Dataset}
We analyze EEG data recorded as 19 subjects listened to Hemingway's
\emph{The Old Man and the Sea}, published in \citet{heilbron2022hierarchy}. The 19 subjects each
listened to the first hour of the recorded story while maintaining
fixation. We analyze 5 sensors distributed
across the centro-parietal scalp: one midline sensor and two lateral sensors per hemisphere at central and posterior positions.
The EEG data were acquired using a 128-channel ActiveTwo system at
a rate of 512 Hz, and down-sampled offline to 128 Hz and re-referenced to the mastoid channels. We
follow the authors' preprocessing method, which includes band-pass filtering the EEG signal between 0.5 and 8 Hz, visual annotation of
bad channels, and removal of eyeblink components via independent component
analysis.\footnote{See \Cref{sec:baseline-model-unstable} for further details on our choice of band-pass filter width.} The dataset also
includes force-aligned annotations for the onsets and durations of both
words and phonemes in these time series.

We generate a predictor time series {\(X_{t}\)} aligned with this EEG
time series (\Cref{sec:features-ts}), ranging from stimulus features (features of the speech
envelope and spectrogram) to sublexical cognitive features (surprisal
and entropy over phonemes). By including these control features in our
models, we can better understand whether or not there is a cognitive and
neural response to words distinct from responses to  their constituent properties
(see \Cref{sec:words-are-privileged} for further discussion).
We generate in addition a set of word-level feature vectors
{\(X_{v} \in \mathbb{R}^{3 \times n_{w}}\)}, consisting of an onset
feature and

\begin{enumerate}
\tightlist
\item
  word surprisal in context, computed with GPT Neo
  2.7B \citep{gpt-neo},\footnote{Preliminary experiments using our baseline model showed that surprisal estimates from GPT Neo 2.7B best explained held-out EEG signals, compared among other sizes of GPT Neo and OpenAI GPT-2 models \citep{radford2019language,brown2020language}.} and
\item
  word unigram log-frequency, from SUBTLEXus 2 \citep{brysbaert2009moving}.
\end{enumerate}

\paragraph{Likelihood estimation}
Our cognitive model requires an estimate of the confusability between
English phonemes (\Cref{eqn:likelihood}). We draw on the experimental data of
\citet{weber2003consonant}, who estimated patterns of confusion in phoneme
recognition within English consonants and vowels by asking
subjects to transcribe spoken syllables. Their raw data
consists of count matrices {\(\psi_{c},\psi_{v}\)} for consonants and
vowels, respectively, where each cell {\(\psi\lbrack ij\rbrack\)}
denotes the number of times an experimental subject transcribed phoneme
{\(j\)} as phoneme {\(i\)}, summing over different phonological contexts
(syllable-initial or -final) and different levels of acoustic noise in the
stimulus presentation. We concatenate this confusion data into a single matrix, imputing a count of 1 for unobserved confusion pairs, and normalize each column
to yield the required conditional probability distributions.

\section{Results}
\label{sec:results}


\newcommand\heldoutTRFvsBaseTRF{\ensuremath{t=4.91, p=0.000113}}
\newcommand\heldoutShiftvsTRF{\ensuremath{t=2.23, p=0.039}}
\newcommand\heldoutStretchvsTRF{\ensuremath{t=5.15, p=6.70 \times 10^{-5}}}
\newcommand\heldoutSurpvsTRF{\ensuremath{t=7.78, p=3.64 \times 10^{-7}}}
\newcommand\heldoutStretchvsSurp{\ensuremath{t=-0.422, p=0.678}}
\newcommand\heldoutStretchvsShift{\ensuremath{t=5.49,p=3.24 \times 10^{-5}}}

\newcommand\berpQuantEdgeOne{64\text{ ms}}
\newcommand\berpQuantEdgeTwo{159\text{ ms}}

\newcommand\surpQuantEdgeOne{1.33\text{ bits}}
\newcommand\surpQuantEdgeTwo{3.71\text{ bits}}

\newcommand\berpSurpAmplitudeTest{\ensuremath{t=-5.23,p=5.71 \times 10^{-5}}}
\newcommand\berpSurpLatencyTest{\ensuremath{t=2.17,p=0.0440}}

\newcommand\berpFigEarlyColor{blue}
\newcommand\berpFigMidColor{orange}
\newcommand\berpFigLateColor{green}

\begin{figure}[t]
    \includegraphics[width=\linewidth]{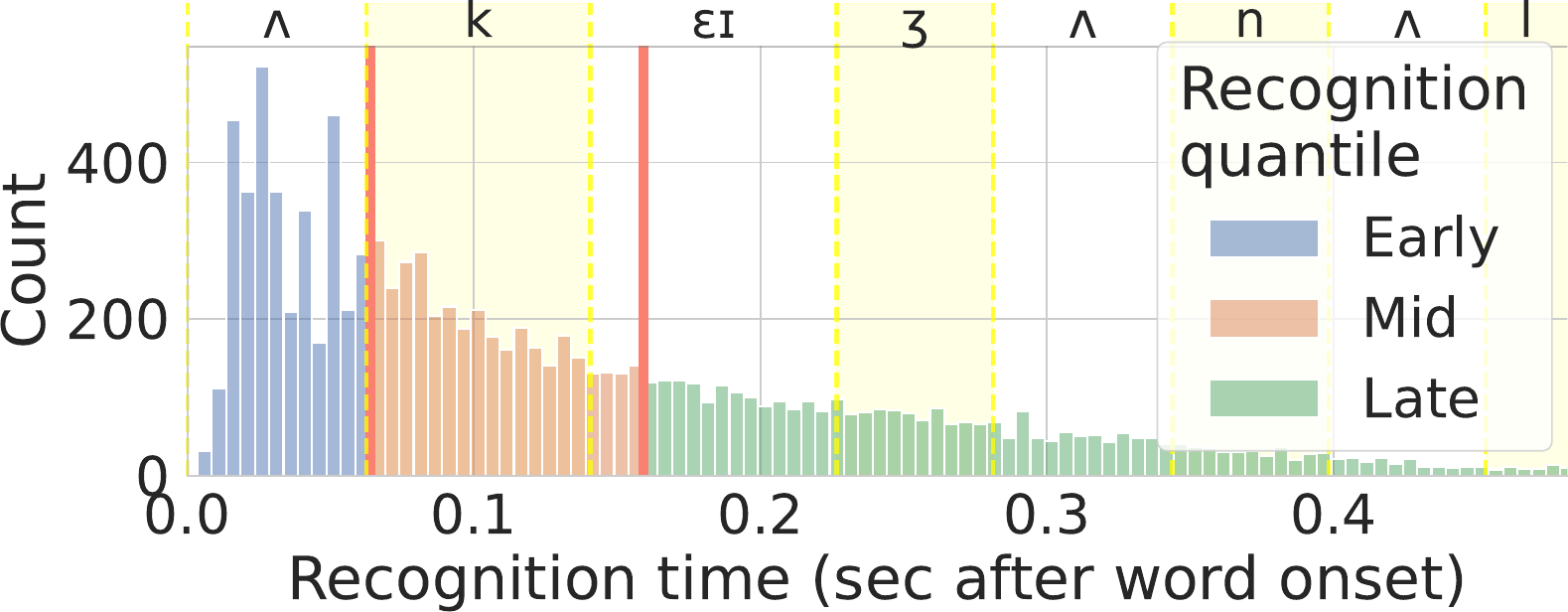}
    \caption{Distribution of inferred recognition times (relative to word onset) for all words, as predicted by the optimal cognitive model parameters. Salmon
vertical lines indicate a tertile partition of words by their recognition time; light yellow regions indicate the median duration of phonemes at each integer position within a word. An example stimulus word, \emph{occasional}, is aligned with phoneme duration regions above the graph.}
    \label{fig:recognition-times}
\end{figure}
\begin{figure*}[t]
    \includegraphics[width=\linewidth]{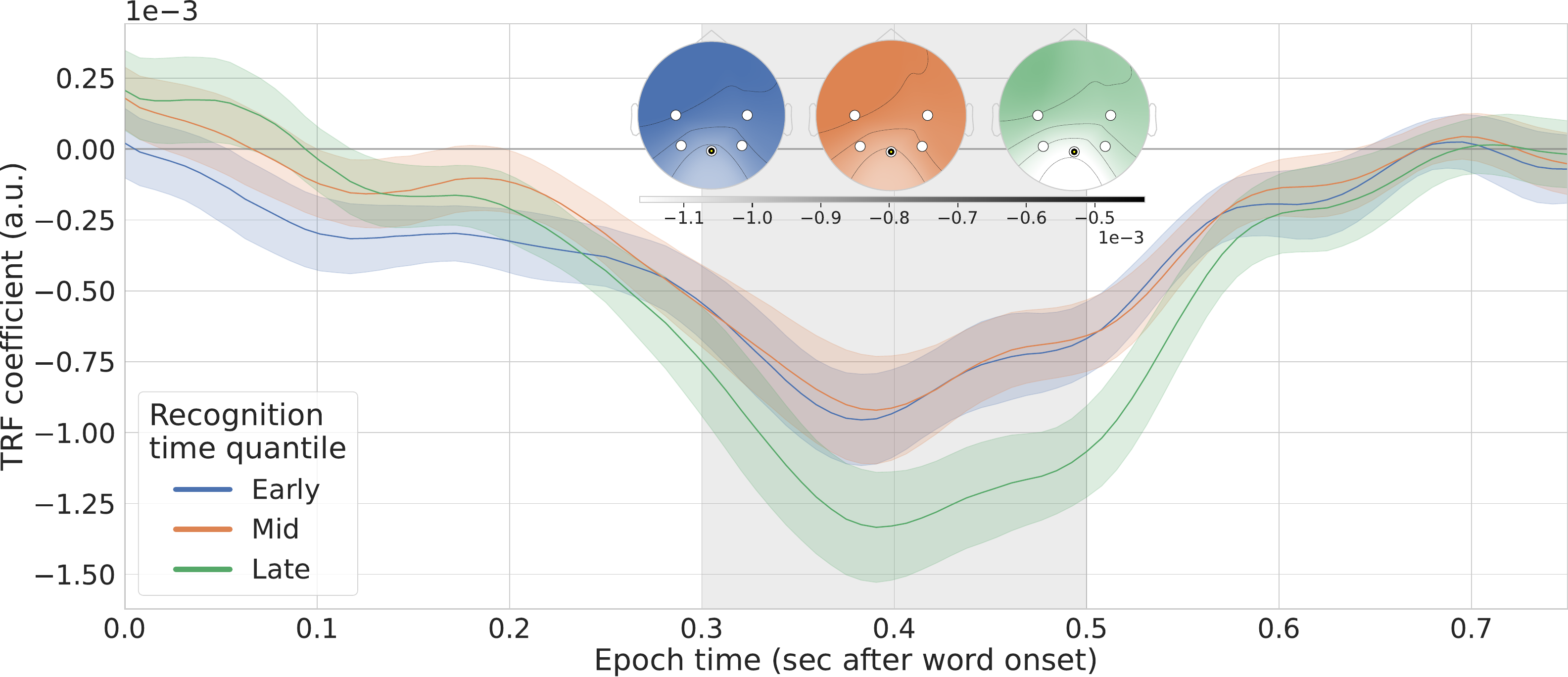}
    \caption{Modulation of scalp voltage at a centro-parietal site by surprisal for words with early ($<\berpQuantEdgeOne{}$, \berpFigEarlyColor{}), middle ($<\berpQuantEdgeTwo{}$, \berpFigMidColor{}), or late ($>\berpQuantEdgeTwo{}$, \berpFigLateColor{}) recognition times. Lines denote inferred coefficients of word surprisal in averaged over subjects for the sensor highlighted in the inset. Error regions denote s.e.m. ($n=19$). Inset: spatial distribution of surprisal modulations averaged for each recognition time quantile within vertical gray regions, where less saturated colors denote more negative response. The surprisal response peaks $\sim$400 ms post onset, amplified for late-recognized words (\berpFigLateColor{}).}
    \label{fig:berp-n400}
\end{figure*}

We first evaluate the baseline model relative to a TRF model which
incorporates no word-level features $X_v$ except for a word onset feature, and find that this
model significantly improves in held-out prediction performance
(\heldoutTRFvsBaseTRF{}). The model recovers a negative response to
word surprisal centered around 400 ms post word onset (\Cref{fig:baseline-n400}),
which aligns with recent EEG studies of naturalistic language
comprehension in both listening \citep{heilbron2022hierarchy,gillis2021neural,donhauser2020two}
and reading \citep{frank2015erp}.

We next separately infer optimal model parameters for the shift and
variable models, and evaluate their error on held-out test data. We find
that the variable model significantly exceeds the baseline model
(\heldoutStretchvsTRF{}), while the shift model does not
(\heldoutShiftvsTRF{}).\footnote{A direct comparison of the variable model and shift model performance also favors the variable model (\heldoutStretchvsShift{}).} This suggests that neural responses to
words are not simply temporally yoked to their recognition times.

We next 
investigate the parameters of the optimal variable model.
\Cref{fig:recognition-times} shows the distribution of predicted word recognition times
{\(\tau_{i}\)} under the optimal variable model on stimulus data from the
held-out test set, charted relative to the onset of a word. Our
model predicts that one third of words are recognized prior to \berpQuantEdgeOne{}
post word onset, another third are recognized between \berpQuantEdgeOne{} and \berpQuantEdgeTwo{},
and a long tail are recognized after \berpQuantEdgeTwo{} post word onset. This
entails that at least a third of words are recognized prior to any
meaningful processing of acoustic input. This prediction aligns with prior work in multiple neuroimaging modalities, which suggests that listeners pre-activate features of lexical items far prior to their acoustic onset in the stimulus \citep{wang2018specific,goldstein2022shared}.

These inferred recognition times maximize the likelihood of the neural
data under the linking variable model parameters {\(\Theta\)}. \Cref{fig:berp-n400}
shows the variable model's parameters describing a neural response to
word surprisal for each of three recognition time quantiles, time locked
to word onset. We see two notable trends in the N400 response which
differ as a function of recognition time:
\begin{enumerate}
\tightlist
\item
  \Cref{fig:berp-n400} shows word surprisal modulations estimated at a centro-parietal site for the three recognition time quantiles.
  Words recognized late (\berpQuantEdgeTwo{} or later post word onset) show an
  exaggerated modulation due to word surprisal. The peak negative
  amplitude of this response is significantly more negative than the
  peak negative response to early words (\cref{fig:berp-n400}, \berpFigLateColor{} line peak minus \berpFigEarlyColor{} line peak in the shaded region;
  within-subject paired \berpSurpAmplitudeTest{}). This modulation is spatially
  distributed similarly to the modulation for early-recognized words
  (compare the \berpFigLateColor{} inset scalp distribution to that of the
  \berpFigEarlyColor{} and \berpFigMidColor{} scalps).
\item
  There is no significant difference in the latency of the N400 response for words
  recognized early vs. late. The time at which the surprisal modulation
  peaks negatively does not significantly differ between early and late words (\cref{fig:berp-n400},
  \berpFigLateColor{}
  line peak time minus \berpFigEarlyColor{} line peak time; within-subject paired
  \berpSurpLatencyTest{}).
\end{enumerate}

These model comparisons and analyses of optimal parameters yield answers to our original questions about the
dynamics of word recognition and integration:
\subparagraph{Response properties:} Neural modulations due to surprisal are exaggerated for words recognized late after their acoustic onset.
\subparagraph{Onset:} The variable model, which asserted integration processes are initiated relative to words' onsets rather than their recognition times, demonstrated a better fit to the data. The optimal parameters under the variable model further showed that while word recognition times seem to affect the amplitude of neural modulations due to surprisal, they do not affect their latency.

\subsection{Prior-variable model}
\label{sec:prior-variable-model}

\begin{figure*}[t]
    \centering
    \begin{subfigure}{0.44\linewidth}
        \raisebox{15pt}{\includegraphics[width=\linewidth]{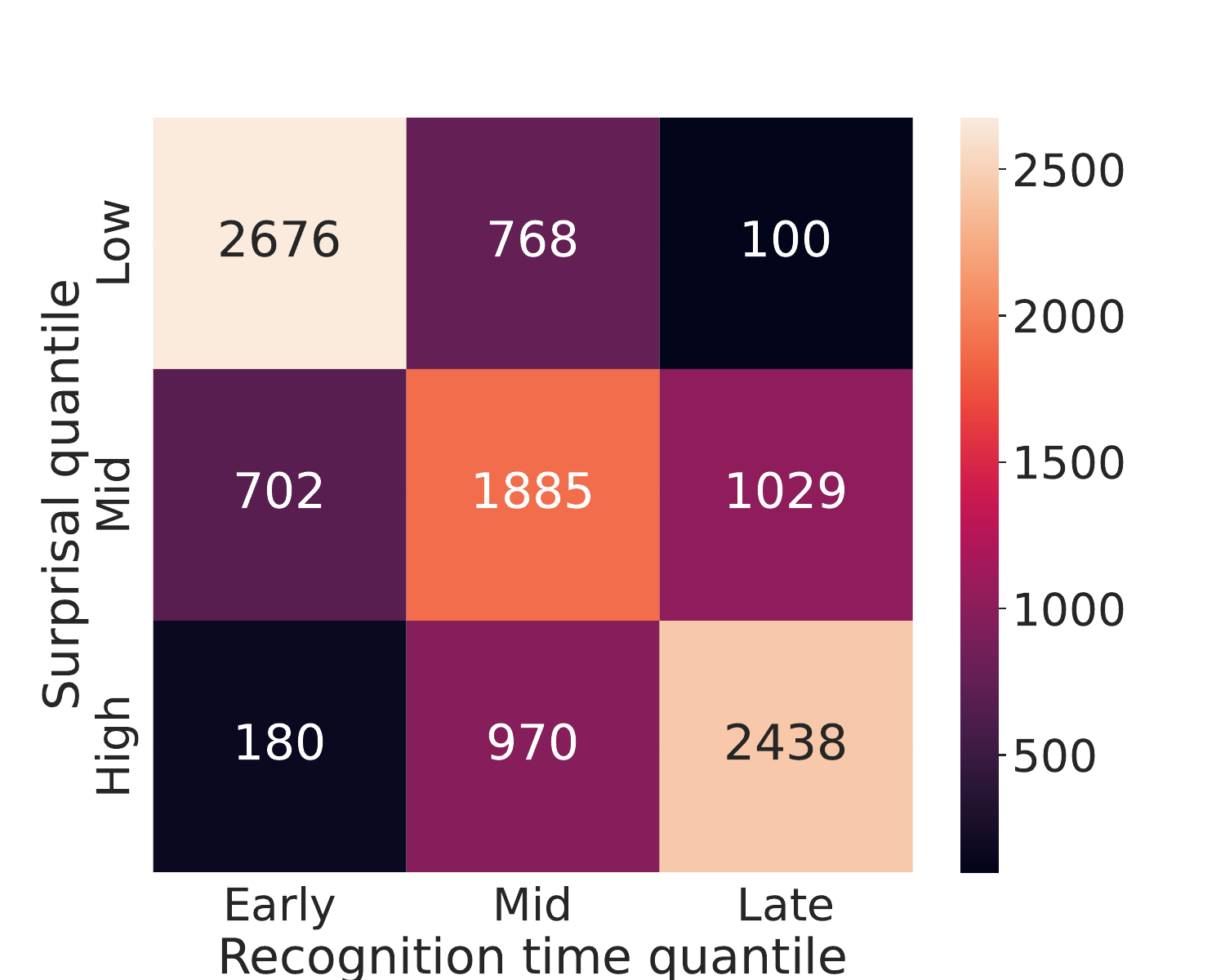}}
        \caption{Confusion matrix comparing partitions of words by the prior-variable model (based on word surprisal; vertical axis) and the optimal word recognition model (based on recognition time; horizontal axis).}
        \label{fig:prior-confusion}
    \end{subfigure}\hfill%
    \begin{subfigure}{0.55\linewidth}
        \begin{subfigure}{\linewidth}
            \resizebox{\linewidth}{!}{
            \begin{tabular}{p{7cm}|p{1.6cm}p{1.6cm}}
                \toprule
                Context & Prior-only prediction & Rec. time prediction \\
                \midrule
                \dots he looked at it in \emph{disgust} & Mid & Late \\
                \dots the old man was now definitely and \emph{finally} & Mid & Late \\
                \dots drew his knife across one of the \emph{strips} & Mid & Late \\
                \midrule
                \dots on his cheeks. The \emph{blotches} & High & Mid \\
                \P He \emph{knelt} & High & Mid \\
                \midrule
                \P ``\emph{I} & Mid & Early \\
                \bottomrule
            \end{tabular}
            }
            \caption{Examples of disagreements in word labeling between the prior-only model and the recognition model.}
            \label{tbl:prior-comparison-qualitative}
        \end{subfigure}
        \begin{subfigure}{\linewidth}
            \centering
            \includegraphics[width=0.5\linewidth]{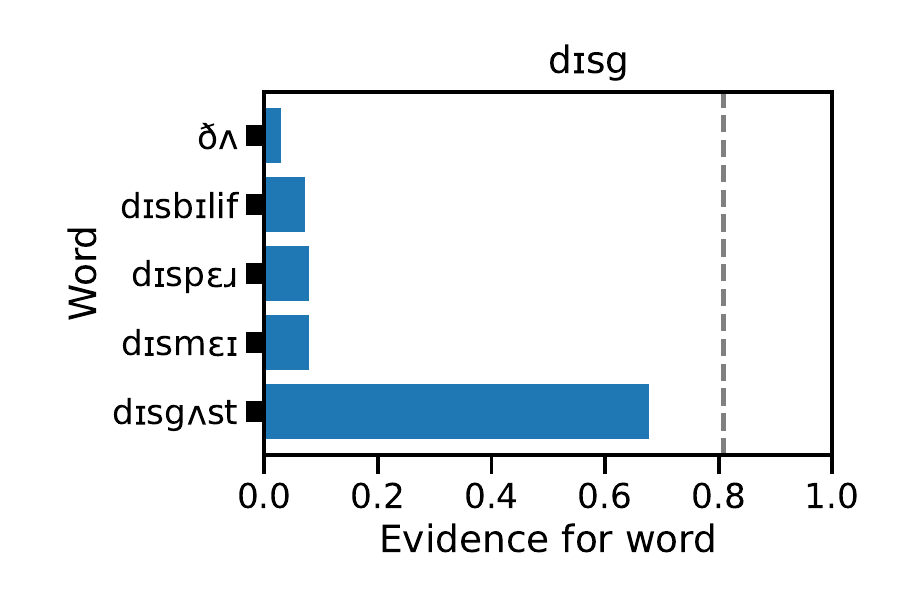}%
            \includegraphics[width=0.5\linewidth]{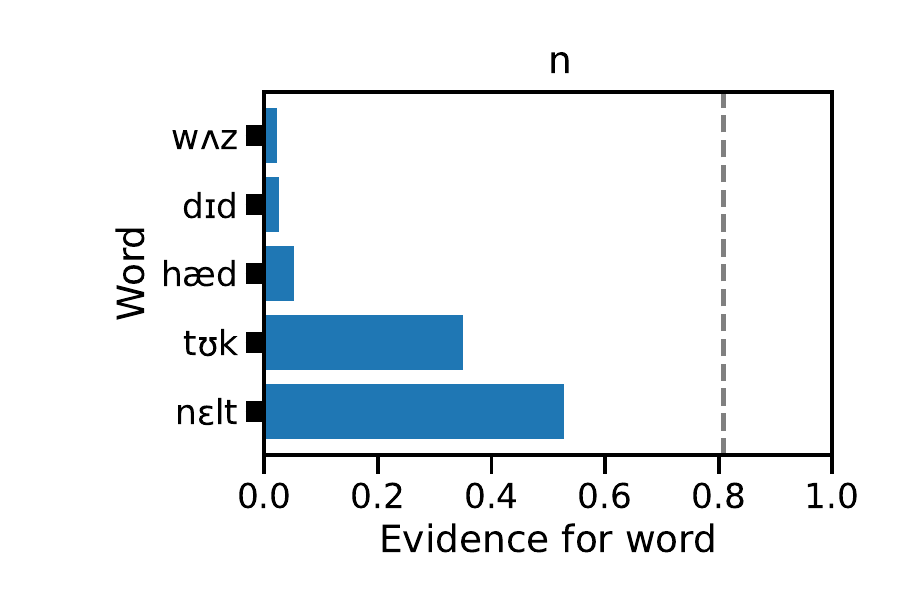}
            \caption{Example posterior predictive distributions for words recognized late due to a dense neighborhood (left); and early due to a sparse neighborhood (right).}
            \label{fig:posteriors}
        \end{subfigure}
    \end{subfigure}
    \caption{Differing predictions of the word recognition model and the prior-variable (surprisal-based) model.}
    \label{fig:model-comparison}
\end{figure*}

We compute a surprisal-based quantile split over words in the training dataset.
The first third of low-surprisal words had a surprisal lower than \surpQuantEdgeOne, while the last third of high-surprisal words had a surprisal greater than \surpQuantEdgeTwo.

We next estimate the prior-variable neural model
parameters, which describe independent neural
responses to words in low\nobreakdash-, mid\nobreakdash-, and high-surprisal quantiles.
This model also significantly exceeds the baseline model (\heldoutSurpvsTRF; see \Cref{sec:prior-variable-results} for inferred model parameters). \Cref{fig:model-comparison} shows a comparison of the way the prior-variable model and the variable model sorted words into different quantiles. While the two models rarely made predictions at the opposite extremes (labeling a low-surprisal word as late-recognized, or a high-surprisal word as early-recognized; bottom left and upper right black corners in \cref{fig:prior-confusion}), there were many disagreements involving sorting words into neighboring time bins (off-diagonal in \cref{fig:prior-confusion}). 
\Cref{tbl:prior-comparison-qualitative,fig:posteriors} show some meaningful cases in which the models disagree to be due to differences in the relevant phonological neighborhood early in the onset of a word. \Cref{fig:posteriors} shows the recognition model's posterior belief over words (\cref{eqn:posterior}) given the incremental phonetic input at the top of the graph. The left panel of \Cref{fig:posteriors} shows how the word \emph{disgust} is recognized relatively late due to a large number of contextually probable phonological neighbors (such as \emph{dismay} and \emph{despair}); the right panel shows how the word \emph{knelt} is recognizable relatively early, since most of the contextually probable completions (\emph{took}, \emph{had}) are likely to be ruled out after the presentation of a second phone.

The variable model's generalization performance is not significantly different than that of this prior-variable model (\heldoutStretchvsSurp{}). Future work will need to leverage other types of neural data to distinguish these models. We discuss this further in \Cref{sec:discussion} and the Limitations section.

\section{Discussion}
\label{sec:discussion}

This paper presented a cognitive model of word recognition which yielded
predictions about the recognition time of words in context $\tau_i$.
A second neural linking model, the variable model, estimated
the neural response to words recognized at early, intermediate, and late
times according to the cognitive model's predictions. This latter model
significantly improved in held-out generalization performance over a
baseline model which did not allow for differences in the neural signal
as a function of a word's recognition time.
We also found, however, that a neural model which
estimated distinct shapes of the neural response to words based on their surprisal --- not their recognition times --- also improved beyond our baseline, and was indistinguishable from the variable model.
More elaborate neural linking theories describing how words' features drive the neural response will be
necessary to distinguish these models \citep[see e.g. the encoding model of][]{goldstein2022shared}.

Our positive findings are consistent with a two-part model of auditory word
recognition and integration, along the lines suggested by 
\citet{vandenbrink2006cascaded} and \citet[\S 3c]{hagoort2008fractionation}. In
this model, listeners continuously combine their expectations with
evidence from sensory input in order to load possible lexical
interpretations of the current acoustic input into a memory buffer. Our
model's prediction of a word's recognition time {\(\tau_{i}\)} measures
the time at which this buffer resolves in a clear lexical inference.

A second integration process reads out the contents of this buffer and
merges them with representations of the linguistic context. Our latency
results show that the timing of this process is independent of a listener's current confidence in
their lexical interpretations, instead time-locked to word onset. This integration process thus exhibits two distinct modes
depending on the listener's buffer contents: one \emph{standard}, in
which the buffer is clearly resolved, and one \emph{exceptional}, in
which the buffer contents are still ambiguous, and additional
inferential or recovery processes must be deployed in order to proceed
with integration. Future work could spell out this distinction mechanistically in
order to explain how buffers in the ``exceptional'' state elicit these distinct
neural responses.

\subsection{What determines integration timing?}
\label{sec:who-sets-the-clock}

Our findings on the stable timing of the naturalistic N400 align with some prior
claims in the experimental ERP literature \citep[\S 5]{federmeier2009time}.\footnote{This is a claim about the \emph{within-subject} consistency of N400 timing, despite substantial between-subject variability, for example, by age and language experience \citep{federmeier2009time}.}
These results strengthen the notion that, even in rapid naturalistic environments,
the timing of the early semantic integration of word meanings is driven not by 
when words are recognized, but rather by the tick of an external clock.

If this integration process is not sensitive to the status of word recognition,
then what drives its dynamics? \citet{federmeier2009time} argue
that this regularly timed integration process is language-external, functioning
to bind early representations of word meaning with existing cognitive representations of the
context via temporal synchrony \citep[see also][]{kutas2011thirty}.
However, other language-internal
mechanisms are also compatible with the data. Listeners may adapt to low-level
features of the stimulus, such as their counterpart's speech rate or prosodic
cues, manipulating the timing of integration to maximize the chances of success in
the expected case.\footnote{See \citet[Figure~6 and Table~4]{Verschueren7442} for evidence against this point, demonstrating that controlled variation in stimulus speech rate does not affect the latency of the N400 response.}

Alternatively, listeners may use the results of the word recognition process to
schedule upcoming attempts at word integration. After recognizing each word
$w_i$, listeners may form an expectation about the likely onset time of word
$w_{i+1}$, using knowledge about the form of $w_i$ and the speech rate.
Listeners could instantiate a clock based on this
prediction, counting down to a time some fixed distance from the expected onset of
$w_{i+1}$, at which semantic integration would be most likely to succeed on
average.
Such a theory could explain how word recognition
and integration are at least approximately optimal given limited cognitive resources \citep{simon1955behavioral,lieder2020resource}: they are
designed to successfully process linguistic inputs in expectation, under the architectural constraint of a fixed
integration clock.

\subsection{Words as privileged units of processing}
\label{sec:words-are-privileged}

Our results suggest that words exist at a privileged level of representation and prediction during speech processing.
This is not a necessary property of language processing: it is possible that word-level processing effects (neural or behavioral responses to word-level surprisal) could emerge as an epiphenomenon of \emph{lower-level} prediction and integration of sublexical units, e.g., graphemes or phonemes. \citet[\S 2.4]{smith2013effect} illustrate how a ``highly incremental'' model which is designed to predict and integrate sublexical units (grapheme- or phoneme-based prediction) but which is measured at higher levels (in word-level reading times or word-level neural responses) could yield apparent contrasts that are suggestive of word-level prediction and integration. On this argument, neural responses to word-level surprisal are not alone decisive evidence for word-level prediction and integration (versus the prediction and integration of sub-lexical units).

Our results add a critical orthogonal piece of evidence in favor of word-level integration: we characterized an integration architecture whose timing is locked to the appearance of word units in the stimulus. While the present results cannot identify the precise control mechanism at play here (\cref{sec:who-sets-the-clock}), the mere fact that words are the target of this timing process indicates an architecture strongly biased toward word-level processing.

\subsection{Prospects for cognitive modeling}

The cognitive model of word recognition introduced in this paper is an extension of Shortlist B \citep{norris2008shortlist}, a race architecture specifying the dynamics of single-word recognition within sentence contexts. We used neural network language models to scale this model to describe naturalistic speech comprehension.
While we focus here on explaining the neural response to words, future work could test the predictions of this model in behavioral measures of the dynamics of word recognition, such as lexical decision tasks \citep{Tucker2018TheMA,ten2022diana}.

\section{Conclusion}

This paper presented a model of the cognitive and neural dynamics of word recognition and integration.
The model recovered the classic N400 integration response, while also detecting a distinct treatment of words based on how and when they are recognized: words not recognized until more than 150 ms after their acoustic onset exhibit significantly amplified neural modulations by surprisal. Despite this processing difference, we found no distinction in the latency of integration depending on a word's recognition time.

However, we developed an alternative model of the neural signal not incorporating word recognition dynamics which also exceeded baseline models describing the N400 integration response. More substantial linking hypotheses bridging between the cognitive state of the word recognition model and the neural signal will be necessary to separate these distinct models.


\section*{Limitations}

There are several important methodological limitations to the analyses in this paper.

We assume for the sake of modeling expediency that all listeners experience the same word recognition dynamics in response to a linguistic stimulus. Individual differences in contextual expectations, attention, and language knowledge certainly modulate this process, and these differences should be accounted for in an elaborated model.

We also assume a relatively low-dimensional neural response to words, principally asserting that the contextual surprisal of a word drives the neural response. This contrasts with other recent brain mapping evaluations which find that high-dimensional word representations also explain brain activation during language comprehension \citep{goldstein2022shared,caucheteux2022brains,schrimpf2021neural}. A more elaborate neural linking model integrating higher-dimensional word representations would likely allow us to capture much more granular detail at the cognitive level, describing how mental representations of words are retrieved and integrated in real time. Such detail may also allow
us to separate the two models (the variable and prior-variable models) which were not empirically distinguished by the results of this paper.


%

\section*{Acknowledgments}
We thank Aixiu An, Jacob Andreas, Canaan Breiss, Trevor Brothers, Tyler Brooke Wilson, Samer Nour Eddine, Evelina Fedorenko, Micha Heilbron, Shailee Jain, Peng Qian, Cory Shain, Jakub Szewczyk, and Josh Tenenbaum for comments on earlier versions of this paper.
We thank Micha Heilbron, Marlies Gillis, and Tamar Regev for invaluable advice on EEG data analysis, and for sharing analysis code and data.
JG gratefully acknowledges support from the Open Philanthropy Project and RPL gratefully acknowledges support from a Newton Brain Science Research Seed Award.

\bibliographystyle{acl_natbib}

\bibliography{main}

\FloatBarrier
\appendix

\section{Relation to pre-activation accounts}
\label{sec:discuss-pre-activation}
Our theoretical account discussed in \Cref{sec:discussion} is partly compatible with \emph{pre-activation}
accounts of prediction in language comprehension, which likewise
suggest that listeners eagerly pre-activate features at multiple levels of 
linguistic representation,
according to both contextual expectations and partial sensory input (see e.g.
\citet{federmeier2007thinking,federmeier2009time,kutas2011thirty,kuperberg2016what} for reviews). 
Our cognitive model of word recognition
provides a mechanism for the temporal dynamics of this
pre-activation process. This mechanism is
an aggressively incremental process, depending on a probabilistic inference
which repeatedly integrates novel acoustic evidence with existing expectations
drawn from the context.


Pre-activation accounts suggest that what is pre-activated are
abstract semantic features rather than specific lexical items 
\citep{federmeier1999rose,kuperberg2016what}. The present model
is stated at the computational level and is thus not directly 
comparable in this respect. Future modeling work can instantiate
specific representational alternatives within this predictive
word recognition model and explore how their predictions might
settle these questions.

\section{Model featurization}
\label{sec:features-ts}

We use a subset of the sublexical features from \citet{heilbron2022hierarchy}
in our TRF models (named as $X_t$ in \Cref{sec:model-neural}).
These features are shared across all models tested in our main
and baseline analysis:
\begin{itemize}
\item
  onset features for each phoneme in the audio stimulus;
\item
  phoneme-onset aligned features:

  \begin{itemize}
  \item
    acoustic control features, averaged within the span of a phoneme:
    average variance in the broadband envelope, and spectral power
    measures averaged within eight bins spaced evenly on a log-mel scale
  \item
    the entropy over a next-phoneme distribution $P(p_j \mid w_{i, <j})$ and the surprisal of the ground-truth phoneme, using the hierarchical predictive model of \citet{heilbron2022hierarchy} (see below).
  \end{itemize}
\end{itemize}

\subsection{Phoneme probability estimator}

The phoneme model of \citet{heilbron2022hierarchy}, whose surprisal and entropy
measures we use as control predictors, combines a word-level language model prior and
a cohort-based likelihood.
For some prior phoneme sequence $p_1,\dots,p_{t-1}$ and some incoming phoneme $p_t$ in
a linguistic context $C$, we define
\begin{align}
    &P(p_t \mid p_1, \dots, p_{t-1}, C) \notag\\
    &\propto \sum_{w \in V} P(w \mid C, p_1, \dots, p_{t-1}) \, P(p_t \mid w) \notag\\
    &= \sum_{w \in V} P(w \mid C) \, \mathbf 1\{ w \in \text{Coh}(p_1, \dots, p_{t-1}, p_t\}
\end{align}
\noindent where $V$ is a vocabulary of all possible word forms, and $\text{Coh}(p_1, \dots, p_t)$ denotes the cohort of a phoneme sequence $p_1, \dots, p_t$ --- i.e., all the words which share the given prefix of phonemes.

This model thus effectively renormalizes a language model's word-level prior $P(w \mid C)$ among words which are exactly phonologically compatible with an observed prefix. See \citet{heilbron2022hierarchy} for further details on the model specification.

\section{Inferred neural response under the prior-variable model}
\label{sec:prior-variable-results}

\begin{figure*}[t]
    \includegraphics[width=\linewidth]{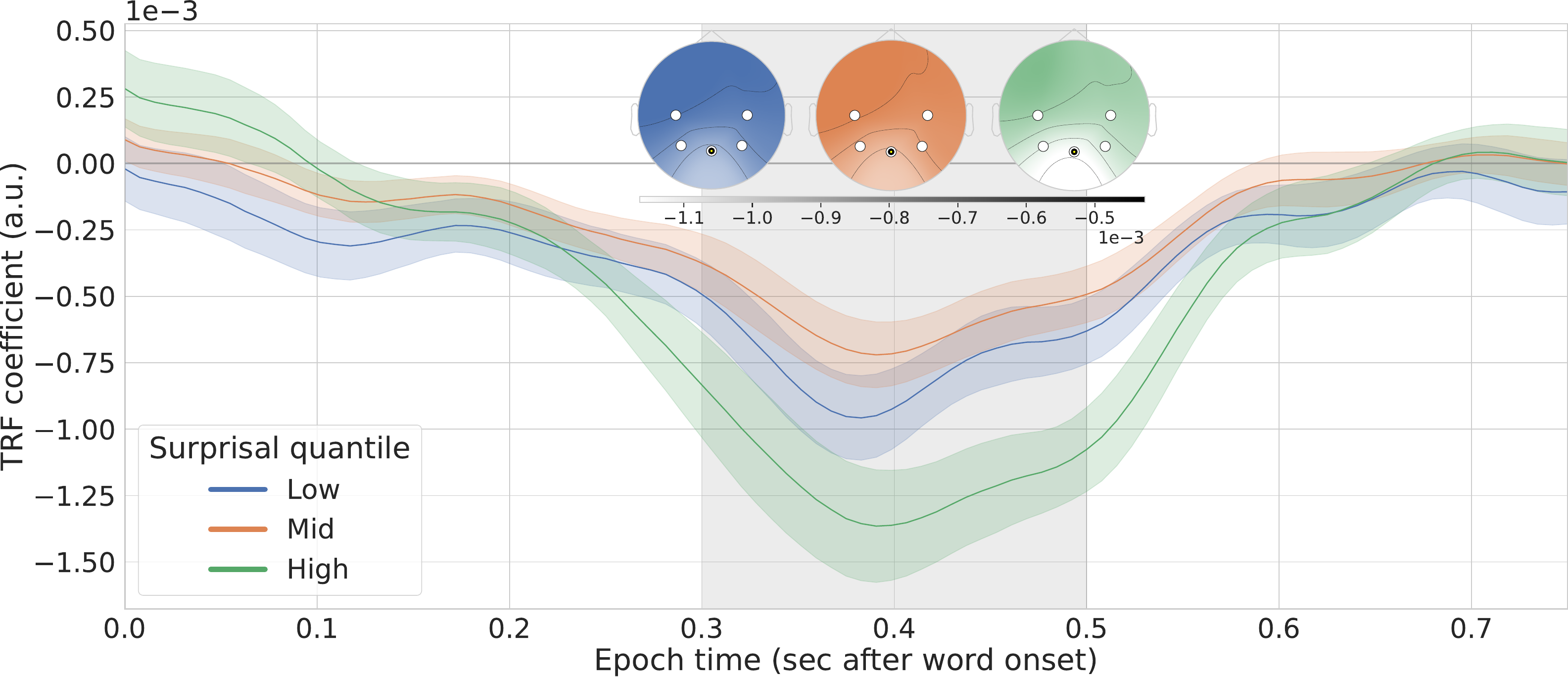}
    \caption{Modulation of scalp voltage at a centro-parietal site by surprisal for words with low ($<\surpQuantEdgeOne$, \berpFigEarlyColor{}), mid ($<\surpQuantEdgeTwo$, \berpFigMidColor{}), or high ($>\surpQuantEdgeTwo$, \berpFigLateColor{}) surprisals. Lines denote inferred coefficients of word surprisal in averaged over subjects for the sensor highlighted in the inset. Error regions denote s.e.m. ($n=19$). Inset: spatial distribution of surprisal modulations averaged for each surprisal quantile within vertical gray regions, where less saturated colors denote more negative response. This is a replication of \Cref{fig:berp-n400} with the parameters of the prior-variable model.}
    \label{fig:prior-variable-n400}
\end{figure*}

\Cref{fig:prior-variable-n400} shows the inferred neural response to words of different surprisal quantiles under the prior-variable model described in \Cref{sec:prior-variable-model}. We see an amplified negative peak in high-surprisal words, similar to that in \Cref{fig:berp-n400} for late-recognized words.

\section{Baseline estimates of the neural response to surprisal}

\begin{figure}[t]
    \includegraphics[width=\linewidth]{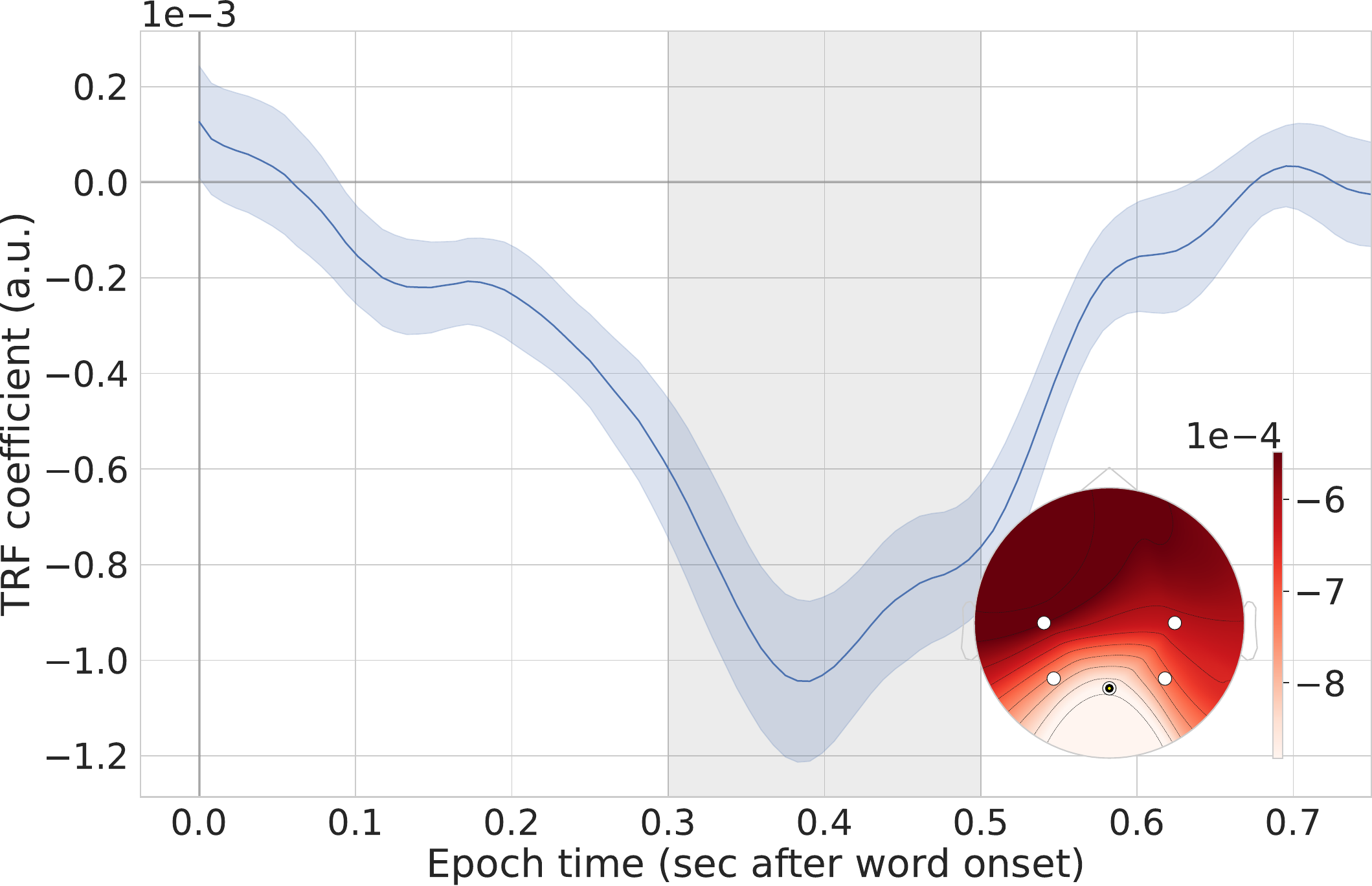}
    \caption{Modulation of scalp voltage by word surprisal in the baseline model at a central posterior sensor, highlighted in inset figure. Error regions denote s.e.m. ($n=19$). Inset: spatial distribution of surprisal modulations averaged within vertical gray region, where less saturated colors denote more negative response.}
    \label{fig:baseline-n400}
\end{figure}

\Cref{fig:baseline-n400} shows the baseline model's estimated response to a word's surprisal. The model recovers the standard broad negative response centered around 400 ms post word onset, which aligns with recent EEG studies of naturalistic language
comprehension in both listening \citep{heilbron2022hierarchy,gillis2021neural,donhauser2020two}
and reading \citep{frank2015erp}.

\begin{figure}[t]
    \includegraphics[width=\linewidth]{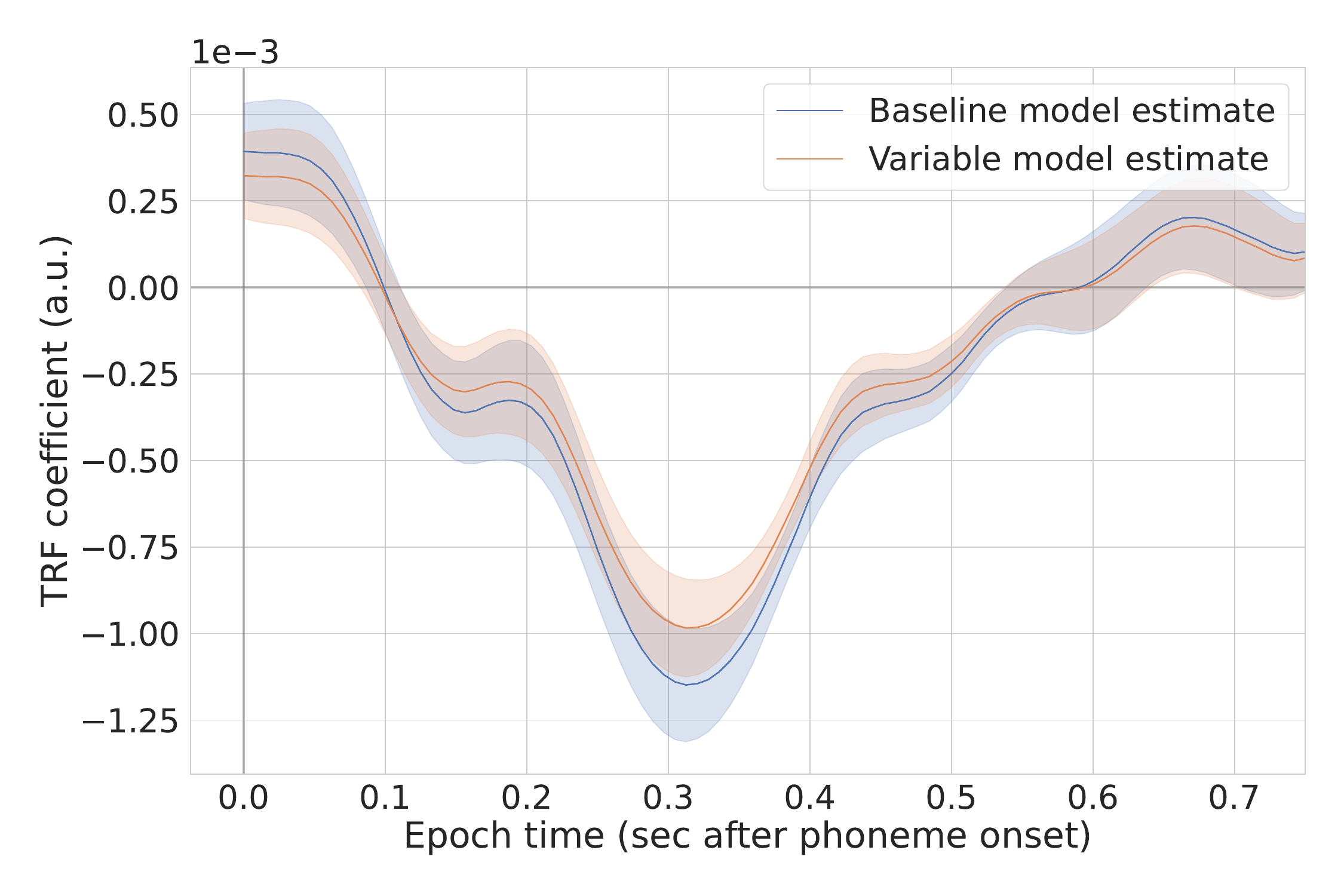}
    \caption{Modulation of scalp voltage at the same central parietal sensor used in \Cref{fig:berp-n400} by phoneme surprisal, estimated in the baseline model and the optimal variable model. Error regions denote s.e.m. ($n=19$).}
    \label{fig:phoneme-surprisal}
\end{figure}

\Cref{fig:phoneme-surprisal} shows estimates of the neural response to phoneme surprisal from both the baseline model and the optimal variable model. All models tested in this paper included this phoneme surprisal predictor; the main results of the paper thus target neural activity above and beyond what is explained by phoneme-level responses. See \Cref{sec:words-are-privileged} for further discussion.

\section{Choice of band-pass filter}
\label{sec:baseline-model-unstable}

\newcommand\stabilityHeldoutStretchvsTRF{}
\newcommand\stabilityBerpSurpAmplitudeTest{\ensuremath{t=-5.03, p=8.71 \times 10^{-5}}}
\newcommand\stabilityBerpSurpLatencyTest{\ensuremath{t=2.17, p=0.043}}
\newcommand\stabilityBerpPSixAmplitudeTest{left central sensor: \ensuremath{t=-0.882, p=0.389}; right central sensor: \ensuremath{t=-1.64, p=0.118}}

A critical preprocessing step in our data analysis is to band-pass filter the raw EEG signal, retaining signals within a frequency window of 0.5--8~Hz. This choice of filter parameters is similar to that of other recent studies of naturalistic language comprehension which use temporal receptive field models \citep[see e.g.][]{gillis2021neural,heilbron2022hierarchy}. A reviewer points out, however, that this filter window is substantially narrower than that of classic controlled studies of the evoked N400 based on trial-averaging ERP analyses \citep[e.g.][]{kutas1984brain,brown1993processing,brothers2023multiple}. This choice of narrow filter parameters for our temporal receptive field analysis has several motivations:
\begin{enumerate}
    \item We wish to focus on evoked responses time-locked to events (e.g. onsets of words and phonemes, and changes in cognitive state due to those stimuli) with rates around this frequency range. Including a wider spectrum adds variance to the signal which we cannot explain using our features of interest,
    \item A high low-cut (more aggressive high-pass filter) allows us to account for signal drift; while this is handled through baselining and detrending in classic ERP analyses, temporal receptive field models have no equivalent capacity to explain drift in the signal.
\end{enumerate}

However, it is possible that this choice of filter parameters could introduce artifacts in the filtered signal which affect the outcomes of our N400-focused analysis. In particular, \citet{tanner2015inappropriate} point out that aggressive high-pass filters ($\sim0.5$ Hz and above) can conflate evoked N400 responses with later ERPs such as the P600, and yield inflated estimates of N400 amplitude.

\subsection{Stability of the baseline model}

\begin{table}
    \centering
    \begin{tabular}{ll|l}
        \toprule
        Low cut & High cut & Result \\
        \midrule
        0.5 Hz & 8 Hz & \heldoutTRFvsBaseTRF{} \\
        0.3 Hz & 8 Hz & $t=3.26, p=0.00435$ \\
        0.1 Hz & 20 Hz & $t=1.95, p=0.0666$ \\
        0.1 Hz & 8 Hz & $t=1.84, p=0.0826$ \\
        \bottomrule
    \end{tabular}
    \caption{Post-hoc stability checks on the baseline model comparison with respect to the low- and high-cut of the band pass filter.}
    \label{tbl:baseline-model-stability}
\end{table}

We thus conducted a post-hoc stability analysis to better understand the sensitivity of this paradigm to our choice of band-pass filter parameters. We first repeated our
initial model comparison on EEG data preprocessed with different band-pass filter parameters. This model comparison evaluates the improvement in predictive performance of a temporal receptive field model which incorporates control acoustic-phonetic features and word-level features (word surprisal and frequency) above a model which does not include these word-level features. (This is the same model comparison described in the beginning of \Cref{sec:results}.) \Cref{tbl:baseline-model-stability} shows the results of this evaluation.

We find that the predictive power of these word-level features diminishes as we decrease the low-cut frequency: beneath 0.3 Hz, this model comparison no longer shows a significant improvement in prediction due to word-level features. We do not take this result to invalidate the claim that word surprisal yields an evoked EEG response in naturalistic comprehension, since this has been supported in other studies of naturalistic comprehension with classic trial-averaging methods \citep{frank2015erp}.

However, it is important to check whether the central finding of this paper --- which rests on an inflated N400 amplitude in response to some types of words --- is sensitive to these parameter changes. In the next section, we reproduce our main qualitative findings for those preprocessing parameters which yield a clear positive baseline outcome of the evoked N400 response to surprisal.

\subsection{Stability of our main findings}

\begin{figure*}[t]
    \includegraphics[width=\linewidth]{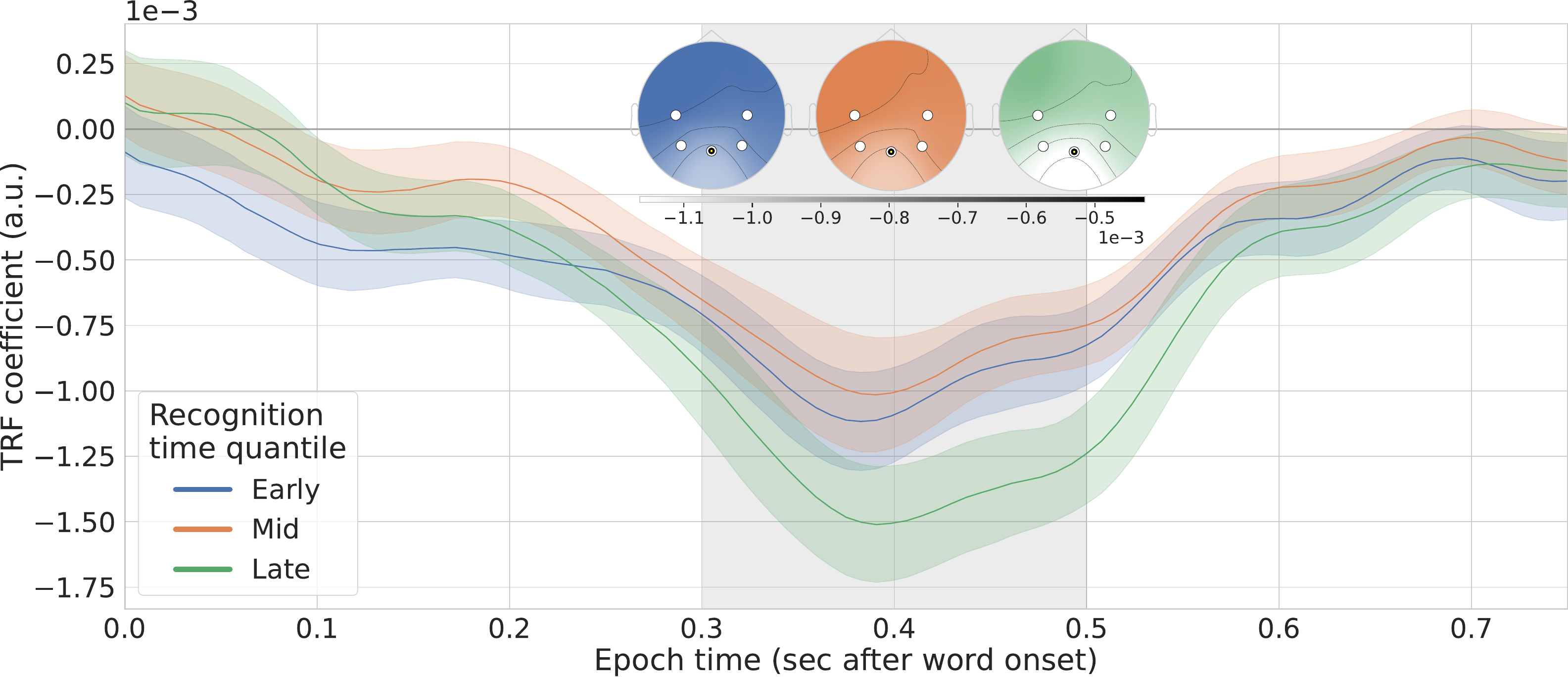}
    \caption{Modulation of scalp voltage at a centro-parietal site by surprisal for words of different recognition time quantiles (according to the variable model of \Cref{fig:berp-n400}), estimated on EEG data band-pass filtered with a low-cut of 0.3~Hz.}
    \label{fig:berp-lowcut-n400}
\end{figure*}

The argument of \citet{tanner2015inappropriate} would predict that the inflated N400 amplitude we observe in response to late-recognized words could be explained away as an artifact of the high-pass filter, which could confound the N400 with a later evoked response (such as the P600). If this finding were purely artifactual, then if we were to relax this high-pass filter, we should see an attenuation of the inflated N400 response and an amplification of a P600 response.

We thus re-fit the temporal receptive field parameters of the optimal variable model described in this paper on EEG data preprocessed with a low-cut of 0.3 Hz, the lowest frequency cut at which the baseline model clearly establishes that an evoked surprisal response is readable in the signal. \Cref{fig:berp-lowcut-n400} shows the estimated neural modulation by word surprisal in these preprocessed data.

We found that this variable model displayed the same qualitative patterns in neural parameters.
Quantitatively, we found a similar effect size of inflated N400 amplitude (\cref{fig:berp-lowcut-n400} \berpFigLateColor{} line peak minus \berpFigEarlyColor{} line peak in the shaded region; within-subject paired \stabilityBerpSurpAmplitudeTest{}).\footnote{Our latency foundings also held null (\berpFigLateColor{}
  line peak time minus \berpFigEarlyColor{} line peak time; within-subject paired \stabilityBerpSurpLatencyTest{}).}

These supplementary analyses suggest that our main findings are stable to different parameterizations of a high-pass filter in EEG preprocessing.

\section{Reproducibility information}

\begin{table*}
    \centering
    \begin{tabular}{l|lp{6cm}}
        \toprule
        (Hyper)Parameter & Bounds & Notes \\
        \midrule
        Regression L2 coefficient & $[10^2, 10^7]$ & Logarithmic space. Bounds manually selected and restricted based on early runs of each model in order to reduce total runtime \\
        \threshold{} (recognition threshold) & $(0, 1)$ & \\
        \temperature{} (evidence temperature) & $[0.1, 3]$ & \\
        \scatterpoint{} (scatter point) & $[0, 1]$ & \\
        \priorscatterpoint{} (prior scatter point) & $[0, 1]$ & \\
        \bottomrule
    \end{tabular}
    \caption{Specifications for parameter and hyperparameter bounds in random search. For details on the meaning of these parameters, see \Cref{tbl:cognitive-parameters}.}
    \label{tbl:hyperparameters}
\end{table*}

\begin{table*}
    \centering
    \begin{tabular}{l|lp{9cm}}
        \toprule
        Model class & Parameter count & Decomposition \\
        \midrule
        Baseline & 138,226 & 138,225 TRF parameters + 1 hyperparameter \\
        Shift & 147,446 & 147,440 TRF parameters + 5 cognitive parameters + 1 hyperparameter \\
        Variable & 230,381 & 230,375 TRF parameters + 5 cognitive parameters + 1 hyperparameter \\
        Prior-variable & 230,375 & 230,374 TRF parameters + 1 hyperparameter \\
        \bottomrule
    \end{tabular}
    \caption{Number of free parameters in all fitted models.}
    \label{tbl:parameter-counts}
\end{table*}

We jointly estimated the parameters of the cognitive model together with the hyperparameters and parameters of the neural linking model using multivariate tree-structured Parzen estimator random search \citep{bergstra2011algorithms} with Optuna \citep{optuna_2019}. For subjects $i=1,\dots,N$, sensors $s=1,\dots,S$, and held-out EEG time series data for subject $i$ at sensor $s$ $Y_{i,s}$, we maximized the value $V$:
\begin{equation}
    V = \frac 1 N \sum_{i=1}^N \left( \max_{s\in \{1,\dots,S\}} r(Y_{i,s}, \hat Y_{i, s}) \right)
\end{equation}
which is the average across subjects of the maximal Pearson correlation of predicted and observed EEG response among all sensors. \Cref{tbl:hyperparameters} shows the precise bounds for each parameter and hyperparameter in this search procedure. We evaluated 20 trials (random settings of parameters) for the baseline model (which only incorporated the L2 coefficient), and 500 trials for all other models. The model results presented in this paper (in visualizations and statistical tests) correspond to the highest-performing outcome of each grid search.

\Cref{tbl:parameter-counts} shows the total count of free parameters under optimization. These counts do not include the parameters of the language model used to compute word surprisal, or the word recognition model likelihood parameters, since these were kept fixed during optimization.

All temporal receptive field models were fit with a receptive field ranging from 0 ms to 750 ms post word onset.


We implemented all training and inference with GPU operations in PyTorch. Due to the large memory requirements of the EEG time series data and the lagged regression computations, we deployed each model fit on two NVIDIA A100 GPUs. Each of the model fits completed in two days or fewer.

\end{document}